%% file: main.tex
\documentclass{article}
\usepackage{times,latexsym}
\usepackage{url}
\usepackage[T1]{fontenc}

\usepackage[]{iclr2022_conference}
\input{package}

\input{macro}
\input{math_commands}

\usepackage{hyperref}
\usepackage{url}
\usepackage{enumitem}

\definecolor{mydarkblue}{rgb}{0,0.08,0.45}
\hypersetup{
 colorlinks=true,
 citecolor=mydarkblue,
 urlcolor=mydarkblue
}

\title{GNN is a Counter?\\ Revisiting GNN for Question Answering \\ }

\author{\textbf{Kuan Wang}$^1$\thanks{Work done during an internship at MSRA}~~, \textbf{Yuyu Zhang} $^1$, \textbf{Diyi Yang}$^{1}$, \textbf{Le Song}$^{3,4}$ \textbf{\&} \textbf{Tao Qin}$^{2}$ \\
$^1$Georgia Institute of Technology \ \ $^2$Microsoft Research Asia\ \ $^3$BioMap \ \ $^4$MBZUAI\\
\texttt{\{kuanwang, yuyu, dyang888\}@gatech.edu} \\
\texttt{lsong@cc.gatech.edu} \\
\texttt{taoqin@microsoft.com}
}

\date{}

\iclrfinalcopy
\begin{document}
\maketitle
\input{0_abstract}

\input{1_intro}

\input{2_analysis}

\input{3_method}

\input{4_exp}

\input{5_related}

\input{6_conclusion}

\input{main.bbl}
\bibliographystyle{iclr2022_conference}

\newpage
\appendix
\input{7_appendix}

\end{document}

%% file: package.tex
\usepackage{mfirstuc,tabulary}

\usepackage{color,xcolor}
\usepackage{epsfig}
\usepackage{graphicx}

\usepackage{adjustbox}
\usepackage{array}
\usepackage{booktabs}
\usepackage{colortbl}
\usepackage{float,wrapfig}
\usepackage{hhline}
\usepackage{multirow}
\usepackage[toc,page]{appendix}

\usepackage{amsmath,amsfonts,amsthm,amssymb}
\usepackage{bm}
\usepackage{nicefrac}
\usepackage{microtype}

\usepackage{changepage}
\usepackage{extramarks}
\usepackage{fancyhdr}
\makeatletter \def\@fancyvbox#1#2{\vbox{#2}} \makeatother
\usepackage{lastpage}
\usepackage{setspace}
\usepackage{soul}
\usepackage{xspace}
\usepackage{indentfirst}

\usepackage{algorithm, algorithmic}
\usepackage{listings}
\usepackage{enumerate}
\usepackage{mathtools}
\usepackage[symbol]{footmisc}

\usepackage{etoolbox}
\makeatletter
\AfterEndEnvironment{algorithm}{\let\@algcomment\relax}
\AtEndEnvironment{algorithm}{\kern2pt\hrule\relax\vskip3pt\@algcomment}
\let\@algcomment\relax
\newcommand\algcomment[1]{\def\@algcomment{\footnotesize#1}}
\renewcommand\fs@ruled{\def\@fs@cfont{\bfseries}\let\@fs@capt\floatc@ruled
  \def\@fs@pre{\hrule height.8pt depth0pt \kern2pt}%
  \def\@fs@post{}%
  \def\@fs@mid{\kern2pt\hrule\kern2pt}%
  \let\@fs@iftopcapt\iftrue}
\makeatother

%% file: macro.tex
\newcommand{\sect}[1]{Section~\ref{#1}}

\newcommand{\fig}[1]{Figure~\ref{#1}}

\newcommand{\tbl}[1]{Table~\ref{#1}}

\newcommand{\myparagraph}[1]{\vspace{-10pt}\paragraph{#1}}

\def\model{Graph Soft Counter\xspace}
\def\modelshort{GSC\xspace}

%% file: math_commands.tex
\usepackage{amsmath,amsfonts,bm}

\def\eqref#1{equation~\ref{#1}}

\def\Algref#1{Algorithm~\ref{#1}}

\def\1{\bm{1}}

\DeclareMathAlphabet{\mathsfit}{\encodingdefault}{\sfdefault}{m}{sl}
\SetMathAlphabet{\mathsfit}{bold}{\encodingdefault}{\sfdefault}{bx}{n}

\def\gE{{\mathcal{E}}}

\def\gG{{\mathcal{G}}}

\def\gR{{\mathcal{R}}}

\def\gV{{\mathcal{V}}}

\newcommand{\mean}{\mathbb{E}}
\newcommand{\var}{{\rm I\kern-.3em D}}

\newcommand{\RR}{\mathbb{R}}
\newcommand{\cond}{\,|\,}
\newcommand{\Normal}{\mathcal{N}}
\newcommand{\diag}{\mathrm{diag}}

%% file: 0_abstract.tex
\begin{abstract}
\label{sec:abstract}

Question Answering (QA) has been a long-standing research topic in AI and NLP fields, and a wealth of studies have been conducted to attempt to equip QA systems with human-level reasoning capability. To approximate the complicated human reasoning process, state-of-the-art QA systems commonly use pre-trained language models (LMs) to access knowledge encoded in LMs together with elaborately designed modules based on Graph Neural Networks (GNNs) to perform reasoning over knowledge graphs (KGs). However, many problems remain open regarding the reasoning functionality of these GNN-based modules. Can these GNN-based modules really perform a complex reasoning process? Are they under- or over-complicated for QA? To open the black box of GNN and investigate these problems, we dissect state-of-the-art GNN modules for QA and analyze their reasoning capability. We discover that even a very simple graph neural counter can outperform all the existing GNN modules on \textit{CommonsenseQA} and \textit{OpenBookQA}, two popular QA benchmark datasets which heavily rely on knowledge-aware reasoning. Our work reveals that existing knowledge-aware GNN modules may only carry out some simple reasoning such as counting. It remains a challenging open problem to build comprehensive reasoning modules for knowledge-powered QA.

\end{abstract}

%% file: 1_intro.tex
\section{Introduction}
\label{sec:intro}

Accessing and reasoning over relevant knowledge is the key to Question Answering (QA).
Such knowledge can be implicitly encoded or explicitly stored in structured knowledge graphs (KGs). 
Large pre-trained language models \citep{devlin2018bert, radford2018improving, radford2019language, brown2020language} are found to be effective in learning broad and rich implicit knowledge \citep{petroni2019language, Bosselut2019COMETCT, talmor2020leap} and thus demonstrate much success for QA tasks. 
Nevertheless, pre-trained LMs struggle a lot with structured reasoning such as handling negation \citep{ribeiro2020beyond, yasunaga2021qagnn}. 
In contrast, the explicit knowledge such as knowledge graphs (KGs) \citep{speer2017conceptnet, bollacker2008freebase} works better for structured reasoning as it explicitly maintains specific information and relations and often produces interpretable results such as reasoning chains \citep{jhamtani2020grc, khot2020qasc, clark2020transformers}.

To utilize both implicit and explicit knowledge for QA, many existing works combine large pre-trained LMs with Graph Neural Networks (GNNs; \citet{scarselli2008graph, kipf2016semi, velivckovic2017graph}), which are shown to achieve prominent QA performance. These approaches commonly follow a two-step paradigm to process KGs: 1) \emph{schema graph grounding} and 2) \emph{graph modeling for inference}. In Step 1, a schema graph is a retrieved sub-graph of KG related to the QA context and grounded on concepts; such sub-graphs include nodes with concept text, edges with relation types, and their adjacency matrix. In Step 2, graph modeling is carried out via an elaborately designed graph-based neural module. For instance, \citet{kagnet-emnlp19} uses GCN-LSTM-HPA which combines graph convolutional networks \citep{kipf2016semi} and LSTM \citep{hochreiter1997long} with a hierarchical path-based attention mechanism for path-based relational graph representation. \citet{feng2020scalable} extends the single-hop message passing of RGCN \citep{schlichtkrull2018modeling} as multi-hop message passing with structured relational attention to obtain the path-level reasoning ability and intractability, while keeping the good scalability of GNN. \citet{yasunaga2021qagnn} uses a LM to encode QA context as a node in the scheme graph and then utilized graph attention networks \citep{velivckovic2017graph} to process the joint graph.

Given that today’s QA systems have become more and more complicated, we would like to revisit those systems and ask several basic questions: Are those GNN-based modules under- or over-complicated for QA? What is the essential role they play in reasoning over knowledge? To answer these questions, we first analyze current state-of-the-art GNN modules for QA and their reasoning capability. Building upon our analysis, we then design a simple yet effective graph-based neural counter that achieves even better QA performance on \textit{CommonsenseQA} and \textit{OpenBookQA}, two popular QA benchmark datasets which heavily rely on knowledge-aware reasoning.

\input{teaser_figure}

In the analysis part, we employ Sparse Variational Dropout (SparseVD; \citet{molchanov2017variational}) as a tool to dissect existing graph network architectures. SparseVD is proposed as a neural model pruning method in the literature, and its effect of model compressing serves as an indicator to figure out which part of the model can be pruned out without loss of accuracy. We apply SparseVD to the inner layers of GNN modules, using their sparse ratio to analyze each layer's contribution to the reasoning process. Surprisingly, we find that those GNN modules are over-parameterized: some layers in GNN can be pruned to a very low sparse ratio, and the initial node embeddings are dispensable.

Based on our observations, we design \model (\modelshort), a very simple graph neural model which basically serves as a counter over the knowledge graph. The hidden dimension of \modelshort layers is only 1, thus each edge/node only has a single number as the hidden embedding for graph-based aggregation. As illustrated in \fig{fig:teaser}, \modelshort is not only very efficient but also interpretable, since the aggregation of 1-dimensional embedding can be viewed as soft counting of edge/node in graphs. Although GSC is designed to be a simplistic model, which has less than 1\% trainable parameters compared to existing GNN modules for QA, it outperforms state-of-the-art GNN counterparts on two popular QA benchmark datasets. Our work reveals that the existing complex GNN modules may just perform some simple reasoning such as counting in knowledge-aware reasoning.

The key contributions of our work are summarized as follows:

\begin{itemize}
	\item \textit{Analysis of existing GNN modules:} We employ SparseVD as a diagnostic tool to analyze the importance of various parts of state-of-the-art knowledge-aware GNN modules. We find that those GNN modules are over-complicated for what they can accomplish in the QA reasoning process.
	\item \textit{Importance of edge counting:} We demonstrate that the counting of edges in the graph plays a crucial role in knowledge-aware reasoning, since our experiments show that even a simple hard counting model can achieve QA performance comparable to state-of-the-art GNN-based methods.
	\item \textit{Design of GSC module:} We propose \model (\modelshort), a simple yet effective neural module as the replacement for existing complex GNN modules. With less than 1\% trainable parameters compared to existing GNN modules for QA, our method even outperforms those complex GNN modules on two benchmark QA datasets.

\end{itemize}

%% file: teaser_figure.tex
\begin{figure}[!t]
    \centering
    \vspace{-14pt}
    \includegraphics[width=\linewidth]{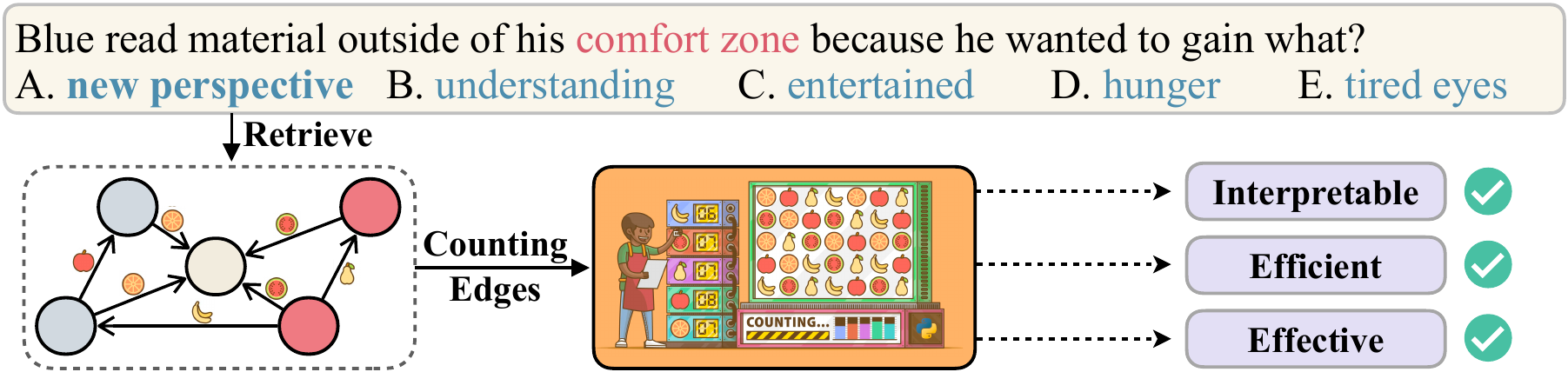}
    \caption{We analyze state-of-the-art GNN modules for the task of KG-powered question answering, and find that the counting of edges in the graph plays an essential role in knowledge-aware reasoning. Accordingly, we design an efficient, effective and interpretable graph neural counter module for knowledge-aware QA reasoning.}
    \label{fig:teaser}
    \vspace{-6pt}
\end{figure}

%% file: 2_analysis.tex
\section{Analysis}
\label{sec:analysis}
\input{svd_overview_figure}

\subsection{Preliminaries}

The knowledge required by QA systems typically comes from two sources: implicit knowledge in pre-trained language models and explicit knowledge in knowledge graphs.

To use the implicit knowledge, existing works commonly use LMs as the encoder to encode textual input sequence $\mathbf{x}$ to contextualized word embeddings, and then pool the embedding of the start token (e.g., $[CLS]$ for BERT) as the sentence embedding. In addition, a MLP (we use a one layer fully connected layer) is used to map the sentence embedding to the score for the choice.

To process the explicit knowledge in knowledge graphs, existing works commonly follow a two-step paradigm: \emph{schema graph grounding} and \emph{graph modeling for inference}. The schema graph is a retrieved sub-graph of KG grounding on concepts related to the QA context. We define the sub-graph as a multi-relational graph $\gG=(\gV, \gE)$, where $\gV$ is the set of entity nodes (concepts) in the KG and $\gE \subseteq$ $\gV \times \gR \times \gV$ is the set of triplet edges that connect nodes in $\gV$ with relation types in $\gR$. Following prior works \citep{kagnet-emnlp19, yasunaga2021qagnn}, we link the entities mentioned in the question $q$ and answer choice $a \in \mathcal{C}$ to the given sub-graph $\gG$.

A wealth of existing works have explored to incorporate pre-trained language models with knowledge-aware graph neural modules. These methods usually have complex architecture design to process complex input knowledge. As shown in \fig{fig:svd_overview}, the processed retrieved sub-graph can be summarized as four main components:

\myparagraph{Node embeddings.} Many existing works on GNN-based QA employ external embeddings to initialize the node embeddings in the graph. For example, \citet{kagnet-emnlp19} employs TransE \citep{wang2014knowledge} with GloVe embeddings \citep{pennington2014glove} as the initial node embeddings. \citet{feng2020scalable} and \citet{yasunaga2021qagnn} use mean-pooled BERT embeddings of entities across the graph as the initialization of node embeddings.

\myparagraph{Relevance score.} In order to measure the quality of a path and prune the sub-graph, \citet{kagnet-emnlp19} decomposes the KG into a set of triples, the relevance score of which can be directly measured by the scoring function of the graph embeddings. \citet{yasunaga2021qagnn} scores the relevance between the retrieved concept and the QA context by the masked LM loss of the stitched sequence. The relevance score determines the priority of the retrieved concepts and also be used as the node score embedding in the input node features. 

\myparagraph{Adjacency matrix.}
The original adjacency matrix of the sub-graph is asymmetric. Typically, the adjacency matrix is converted to be symmetric before feeding into the GNN module: each KG relation is double recorded with opposite directions as symmetric edges in the processed adjacency matrix.

\myparagraph{Edge embeddings.}
Existing works use various ways to obtain the edge embeddings. \citet{yasunaga2021qagnn} uses concatenated one-hot vector $[u_s, e_{st}, u_t]$ to encode edge triplets, where $u_s, u_t$ indicates the node type and $e_{st}$ indicates the edge type. \citet{kagnet-emnlp19} uses TransE \citep{wang2014knowledge} initialized with GloVe embeddings to generate edge embeddings. \citet{feng2020scalable} designs product-based edge embeddings to approximate the multi-hop relations in the graph.

\input{svd_curve_figure}

\subsection{Dissection}

To investigate the mechanism of these complex systems and how they use the complex information, we introduce a neural model pruning method named Sparse Variational Dropout (SparseVD) \citep{molchanov2017variational} as a diagnostic tool to automatically dissect the graph network architecture. Note that our dissection tool is pruning method agnostic; other pruning schemes \citep{han2015learning, he2017channel, liu2017learning} may also be applicable. Here we choose SparseVD since it prunes not only the weights with smaller scale but also the weights with higher variance, and it is theoretically supported by stochastic variational inference \citep{hoffman2013stochastic, kingma2013auto, rezende2014stochastic, kingma2015variational}.

\input{svd_table}

Despite the fact that SparseVD was originally proposed in the field of model compression \citep{Han:2016uf, He:2018vj, Lin:2017ww, zhou2018explicit, wang2018haq}, we use it to investigate which parts of GNN can be pruned out (sparse ratio to zero) without loss of accuracy, which indicates that part of the model is redundant. To be specific, we keep the target model architecture unchanged, and parameterize each part of the weights in the model as a Gaussian distribution. Afterwards, this probabilistic model will be trained with a cross-entropy loss jointly with a KL-divergence regularization term. So the joint loss constrains the weights to our pruning prior. We implement the SparseVD with the default threshold as in \citet{molchanov2017variational}. Eventually, we get the pruned model with different sparsified ratios for the layers. As shown in \tbl{tab:svd}, we investigate three representative GNN-based QA methods, and the sparsified models could achieve the accuracy of their original counterparts, which indicates that our dissection regularization does not hurt the models, so that we can dissect into each layer to see what helps the model to do reasoning.

\input{gsc_overview_figure}

\subsection{Observations and Hypothesis}

In \fig{fig:svd_curve}, we plot the sparse ratio of some representative layers during the SparseVD training of GNN reasoning modules. Due to the limited space, refer to Appendix~\ref{sec:appendix_a} for the full plots of curves. According to thees plots, we summarize our key observations as follows: 1) the left plot shows that the edge encoder that encodes edge triplets (edge/node type information) preserves a relatively higher sparse ratio, while the node score embedding layer can be fully pruned; 2) the middle plot shows the layers inside GNN, and all the sparse ratios are low while the value layer has a relatively higher sparse ratio than key/query layers; 3) the right plot shows the concept embedding layers of three representative GNN methods, which process the initial node embeddings, can be completely discarded. Inspired by these findings, we come up with the following design guidelines for a much simpler yet effective graph neural module for knowledge-aware reasoning:
\myparagraph{Node embeddings.} The initial node embedding and the extra score embedding are shown to be dispensable in these scenarios, thus we can directly remove the embedding layers. 

\myparagraph{Edge embeddings.} The edge encoder layers are hard to prune which indicates that edge/node type information are essential to reasoning, so we further leverage it to get a better representation.
\input{gsc_algorithm}

\myparagraph{Message passing layers.} The linear layers inside GNN layers (query, key, value, etc) can be pruned to a very low sparse ratio, suggesting that GNN may be over-parameterized in these systems, and we can use fewer parameters for these layers. 

\myparagraph{Graph pooler.}  The final attention-based graph pooler aggregates the node representation over the graph to get a graph representation. We observe that the key/query layers inside the pooler can be pruned out; as a result, the graph pooler can be reduced to a linear transformation.

%% file: svd_overview_figure.tex
\begin{figure}[!t]
    \centering
    \vspace{-20pt}
    \includegraphics[width=0.95\linewidth]{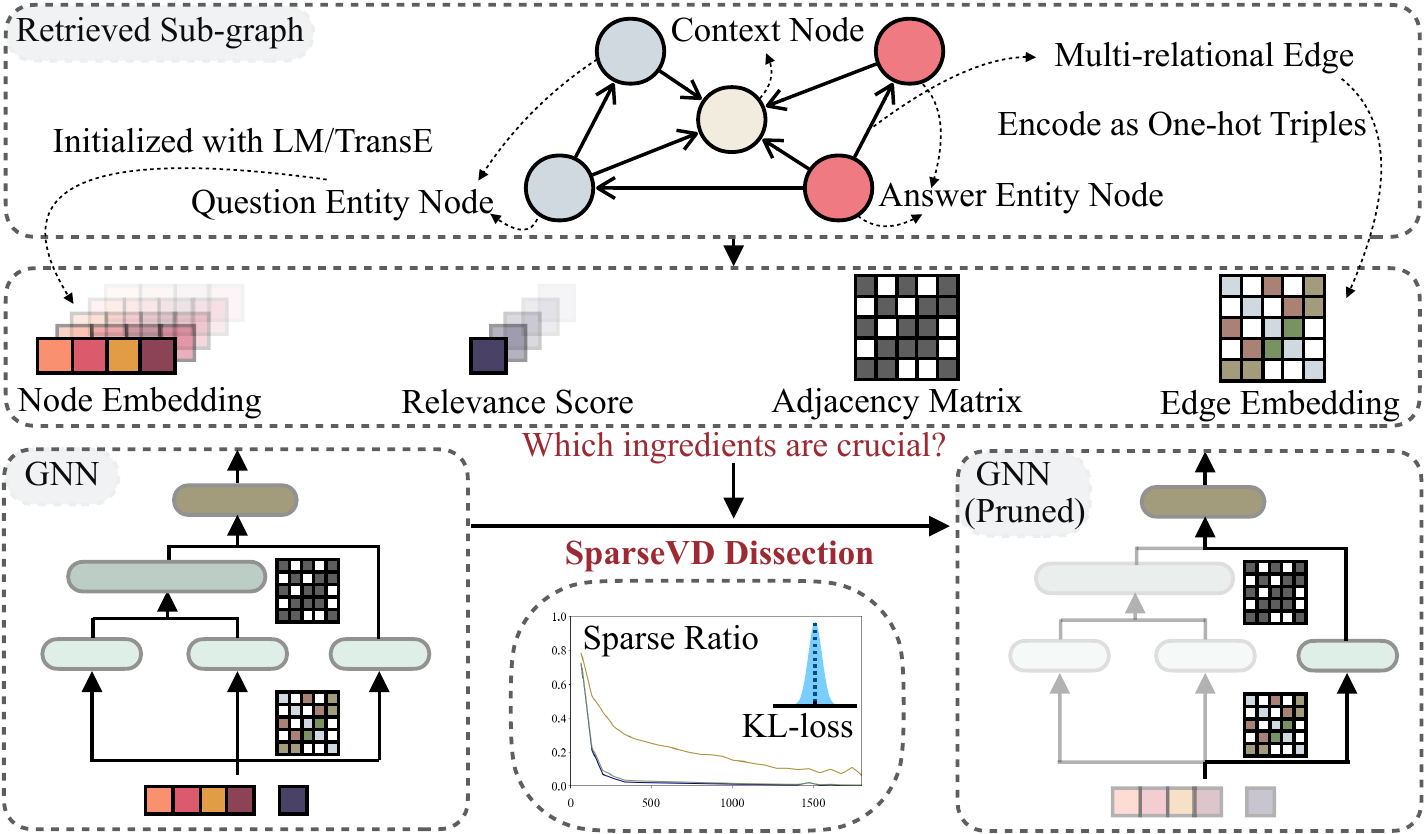}
    \caption{The retrieved sub-graph of KG is formulated as entity nodes representing concepts connected by edges representing relations and the central context node connected to all question entity nodes and answer entity nodes. The pre-processed graph data generally has the following ingredients: node embedding initialized with pre-trained KG embeddings, relevance score computed by LM, adjacency matrix representing the topological graph structure, edge embeddings to encode the node types, and edge information of relation types. We adapt SparseVD as a diagnostic tool to dissect GNN-based reasoning modules for QA, getting the sparse ratio of each layer to indicate its importance. We find that some layers and ingredients are completely dispensable, which inspires us to design a simple, efficient and effective GNN module as the replacement of existing complex GNN modules.}

    \label{fig:svd_overview}
    \vspace{-14pt}
\end{figure}

%% file: svd_curve_figure.tex
\begin{figure*}[t]
    \centering
    \vspace{-28pt}
    \includegraphics[width=\linewidth]{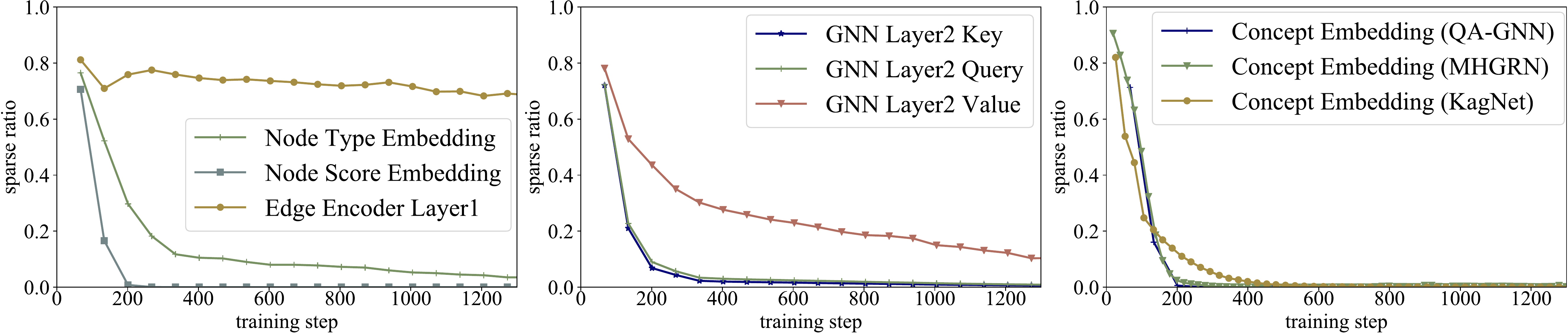}
    \vspace{-18pt}
    \caption{The sparse ratio curves obtained from the SparseVD training of GNN reasoning modules: 1) the left plot shows the curves for the embedding layers in QA-GNN, where the node score and the node type reach a low ratio, while the edge encoder preserves a large ratio; 2) the middle plot shows the curves for the layers within the QA-GNN graph attention layer, where the key and query layers converge to a fairly low ratio, while the value layer has a relatively large ratio; 3) the right plot shows the curves of the initial node embedding layers of three representative GNN-based QA methods, where all the ratios are close to 0, indicating that the initial node embeddings are dispensable.}
    \label{fig:svd_curve}
    \vspace{-8pt}
\end{figure*}

%% file: svd_table.tex
\begin{table}[b]
\vspace{-14pt}
\renewcommand*{\arraystretch}{0.1}

\scalebox{0.92}{\parbox{\linewidth}{
\begin{tabular}{lcc|cc}
    \toprule  
    & \multicolumn{2}{c}{w/o SparseVD} & \multicolumn{2}{c}{w/ SparseVD} \\
    \cmidrule(lr){2-3}\cmidrule(lr){4-5} 
    {\textbf{Methods}}& \textbf{IHdev-Acc.} (\%) & \textbf{IHtest-Acc.} (\%) & \textbf{IHdev-Acc.} (\%) & \textbf{IHtest-Acc.} (\%)  \\
    \midrule
    KagNet \scalebox{0.7}{\citep{kagnet-emnlp19}} &  73.47~($\pm$0.22)   & 69.01~($\pm$0.76) &  75.18~($\pm$1.05)   & 70.48~($\pm$0.77) \\
    \midrule
    MHGRN \scalebox{0.7}{\citep{feng2020scalable}} &  74.45~($\pm$0.10)   & 71.11~($\pm$0.81) &  77.15~($\pm$0.32)   & 72.66~($\pm$0.61) \\
    \midrule
    QAGNN \scalebox{0.7}{\citep{yasunaga2021qagnn}} &  76.54~($\pm$0.21)   & 73.41~($\pm$0.92) &  77.64~($\pm$0.50)   & 73.57~($\pm$0.48) \\
    
    \bottomrule 
\end{tabular}
}}
\vspace{-4pt}
\caption{To preserve the reasoning ability for analysis, our SparseVD tool prunes the GNN-based models without loss of accuracy on \textit{Commonsense \!QA} in-house split. 
}
\vspace{-8pt}
\label{tab:svd}
\end{table}

%% file: gsc_overview_figure.tex
\begin{figure}[t]
    \vspace{-14pt}
    \centering
    \includegraphics[width=0.95\linewidth]{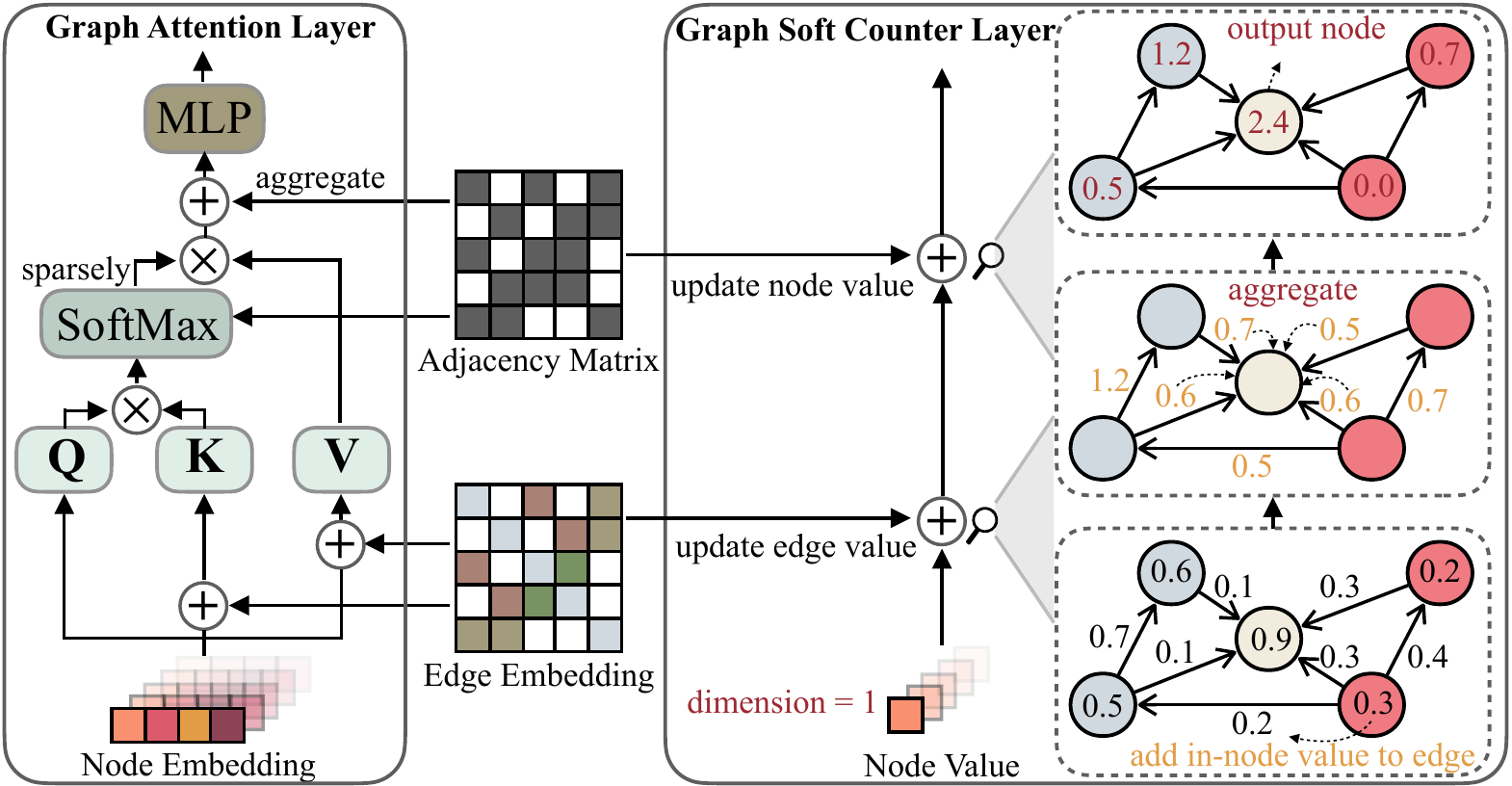}
    \caption{\model (right) extremely simplifies the architecture of conventional GNNs (left). The \modelshort layers are parameter-free and only keep the basic graph operations: 1) update each edge embedding with incoming node (in-node) embeddings; 2) update each node embedding by aggregating the edge embeddings. Since we reduce the hidden dimension to only 1, GSC can be viewed as a soft counter over the graph for generating the final score of the output node.}
    \vspace{-14pt}
    \label{fig:gsc_overview}
\end{figure}

%% file: gsc_algorithm.tex
\begin{wrapfigure}{R}{.55\textwidth}
\begin{minipage}{.55\textwidth}
\vspace{-8mm}
\begin{algorithm}[H]

\definecolor{codeblue}{rgb}{0.25,0.5,0.5}
\lstset{
  backgroundcolor=\color{white},
  basicstyle=\fontsize{7.2pt}{7.2pt}\ttfamily\selectfont,
  columns=fullflexible,
  breaklines=true,
  captionpos=b,
  commentstyle=\fontsize{7.2pt}{7.2pt}\color{codeblue},
  keywordstyle=\fontsize{7.2pt}{7.2pt},
}
\begin{lstlisting}[language=python]
# qa_context: question answer pair context
# adj: edge index with shape 2 x N_edge
# edge_type: edge type with shape 1 x N_edge
# node_type: node type with shape 1 x N_node
edge_emb = edge_encoder(adj, edge_type, node_type)
node_emb = torch.zeros(N_node)
for i_layer in range(num_gsc_layers):
    # propagate from node to edge
    edge_emb += node_emb[adj[1]]
    # aggregate from edge to node 
    node_emb = scatter(inputs, adj[0])
graph_score = node_emb[0]
context_score = fc(roberta(qa_context))
qa_score = context_score + graph_score
\end{lstlisting}
\caption{PyTorch-style code of GSC}
\label{alg:code}
\end{algorithm}
\end{minipage}
\vspace{-14pt}
\end{wrapfigure}

%% file: 3_method.tex
\section{\model}
\label{sec:method}

Based on our findings in \sect{sec:analysis}, we design the \model (\modelshort), a simplistic graph neural module to process the graph information. As demonstrated in \Algref{alg:code}, there are only two basic components in \modelshort to compute the graph score: Edge encoder and \model layers. We can get the QA choice score by simply summing up the graph score and context score.

\input{method_compare_table}

The edge encoder is a two-layer MLP with dimension $[47\times32\times1]$ followed by a Sigmoid function to encode the edge triplets to a float number in the range of $(0, 1)$. The triplets is represented as a concatenated one-hot vector $[u_s, e_{st}, u_t]$, where $u_s, u_t$ indicates 4 node types and $e_{st}$ indicates 38 relation types (17 regular relations plus question/answer entity relations and their reversions). 

\model layers are parameter-free and only keep the basic graph operations: Propagation and aggregation. To overcome over-parameterization, one straightforward way is to reduce the hidden size, and the extreme case is reducing it to only 1. As shown in \fig{fig:gsc_overview}, the \modelshort layer simply propagates and aggregates the numbers on the edges and node following the two-step scheme: 1) update the edge value with in-node value, and this is done by simply indexing and adding; 2) update the node value by aggregating the edge values, and this is done by scattering edge values to out-node.

Since we reduce the hidden size to only 1, there are basically numbers propagating and aggregating on the graph. We can interpret these numbers as soft counts representing the importance of the edges and nodes. The aggregation of \modelshort can be regarded as accumulating the soft counts, so call it \model. In addition, we can also formulate \modelshort as an extremely efficient variant of current mainstream GNN such as GAT and GCN. As shown in \tbl{tab:complexity}, the computation complexity of \modelshort is much smaller than the baselines regard to both time and space. \tbl{tab:method_compare} showing the trainable parameters in \modelshort is remarkably less than previous methods since our message-passing layers are parameter-free. The model size of \modelshort is also extremely small since we do not use any initial node embedding. 

To draw a more conclusive conclusion, we handcraft a counting-based feature and use a simple two-layer MLP to map it to graph score. The feature is computed by counting the occurrence of each possible edge triplet. Surprisingly, this simple model could achieve similar performance and can even outperform baselines, which further proves that Counter is a basic and crucial part in QA. 

%% file: method_compare_table.tex
\begin{wraptable}{r}{0.5\textwidth}
\scalebox{0.68}{\parbox{\linewidth}{
\begin{tabular}{lccc|c}
    \toprule  

    & KagNet & MHGRN & QAGNN & \textbf{GSC} (Ours) \\
    \midrule  
    \textbf{Adj-matrix}      & \checkmark & \checkmark & \checkmark & \checkmark  \\
    \textbf{Edge-type}       & \checkmark & \checkmark & \checkmark & \checkmark  \\
    \textbf{Node-type}       & $\times$   & \checkmark & \checkmark & \checkmark  \\
    \textbf{Node-embedding}  & \checkmark & \checkmark & \checkmark & $\times$    \\
    \textbf{Relevance-score} & $\times$   & $\times$   & \checkmark & $\times$    \\
    \midrule  
    \textbf{\#Learnable Param}     & 700k       & 547k       & 2845k      & \textbf{3k} \\
    \midrule  
    \textbf{Model size}      & 819M       & 819M       & 821M       & \textbf{3k} \\
    \bottomrule 
\end{tabular}
}}
\vspace{-6pt}
\caption{ Our \modelshort do not use node embedding and node score versus the previous works, which makes our model size is extremely small (without counting LM). And our model do not have learnable parameter except the edge encoder, so our method has much less chance to be over-parameterized. }
\vspace{10pt}
\label{tab:method_compare}

 \scalebox{0.75}{
 \begin{tabular}{lcc}
  \toprule
  \textbf{Model} & \textbf{Time}  & \textbf{Space}  \\
  \midrule
  \multicolumn{3}{c}{\textit{$\mathcal{G}$ is a dense graph}} \\
  \midrule
  $L$-hop KagNet & \!\!$\mathcal{O}\left(|\gR|^L |\gV|^{L+1}L \right)$\!\! & \!\!$\mathcal{O}\left(|\gR|^L |\gV|^{L+1}L \cdot D \right)$\!\! \\
  $L$-hop MHGRN & $\mathcal{O}\left(|\gR|^2|\gV|^2L \right)$ &  $\mathcal{O}\left ( |\gR||\gV|L \cdot D \right)$ \\
  $L$-layer QAGNN\!\! & $\mathcal{O}\left ( |\gV|^2L \right)$ & $\mathcal{O}\left( |\gR||\gV|L \cdot D \right)$ \\
  $L$-layer \modelshort\!\! & $\mathcal{O}\left ( |\gV|L \right)$ & $\mathcal{O}\left( |\gR||\gV|L \right)$  \\
  \midrule
  \multicolumn{3}{c}{\textit{$\mathcal{G}$ is a sparse graph with maximum node degree $\Delta \ll |\gV|$}} \\
  \midrule
  $L$-hop KagNet & \!\!$\mathcal{O}\left(|\gR|^L|\gV|L\Delta^{L} \right)$\!\! &  \!\!$\mathcal{O}\left(|\gR|^L|\gV|L\Delta^{L} \cdot D \right)$\!\! \\
  $L$-hop MHGRN & $\mathcal{O}\left(|\gR|^2|\gV| L\Delta \right)$ &  $\mathcal{O}\left ( |\gR||\gV|L \cdot D \right)$ \\
  $L$-layer QAGNN \!\! & $\mathcal{O}\left(|\gV|L\Delta \right)$ & $\mathcal{O}\left ( |\gR||\gV|L \cdot D \right)$ \\
  $L$-layer \modelshort \!\! & $\mathcal{O}\left(|\gV|L \right)$ & $\mathcal{O}\left ( |\gR||\gV|L \right)$ \\
\bottomrule
\end{tabular}
}
\vspace{-6pt}
\caption{\modelshort is extremely efficient compared to the computation complexity of $L$-hop reasoning models with hidden dimension $D$ on a dense \!/\! sparse graph $\gG=(\gV,\gE)$ with the relation set $\gR$. 
} 
\vspace{-14pt}
\label{tab:complexity}

\end{wraptable}

%% file: 4_exp.tex
\section{Experiments}
\label{sec:exp}

\subsection{Settings}

\myparagraph{Datasets and KG.} We conduct extensive experiments on \textit{CommonsenseQA} \citep{talmor2018commonsenseqa} and \textit{OpenBookQA} \citep{OpenBookQA2018}, two popular QA benchmark datasets that heavily rely on knowledge-aware reasoning capability. \textit{CommonsenseQA} is a 5-way multiple choice QA task that requires reasoning with commonsense knowledge, which contains 12,102 questions. The test set of \textit{CommonsenseQA} is not publicly available, and the model predictions can only be evaluated once every two weeks via the official leaderboard. Hence, following the data split of \citet{kagnet-emnlp19}, we experiment and report the accuracy on the in-house dev (IHdev) and test (IHtest) splits. We also report the accuracy of our final system on the official test set. \textit{OpenBookQA} is a 4-way multiple choice QA task that requires reasoning with elementary science knowledge, which contains 5,957 questions. Following \citet{clark2019f}, methods with AristoRoBERTa use textual evidence as an external input of the QA context. In addition, we use \textit{ConceptNet} \citep{speer2017conceptnet}, a general commonsense knowledge graph, as our structured knowledge source $\mathcal{G}$ for both of the above tasks. Given each QA context (question and answer choice), we retrieve the sub-graph $\mathcal{G}_\text{sub}$ from $\mathcal{G}$ following \citet{feng2020scalable}. We only use the concepts that occur in the QA context, as detailed in \tbl{tab:ablation}.

\myparagraph{Implementation and training details.}
\input{csqa_main_table}

We use a two-layer MLP with hidden dimension $47\times32\times1$ followed by a Sigmoid function as our edge encoder, and the number of \modelshort layers is 2. Since GSC is extremely efficient, we set the dropout \citep{srivastava2014dropout} rate to 0. We use RAdam~\citep{liu2019radam} as the optimizer and set the batch size to 128. The learning rate is 1e-5 for RoBERTa and 1e-2 for \modelshort. The maximum number of epoch is set to 30 for \textit{CommonsenseQA} and 75 for \textit{OpenBookQA}. On a single Quadro RTX6000 GPU, each GSC training only takes about 2 hours to converge, while other methods often take 10+ hours.

\myparagraph{Baselines.} 
\input{csqa_leaderboard_table}

For GSC and all the other methods, RoBERTa-large \citep{liu2019jingfei} is used for both \textit{CommonsenseQA} and \textit{OpenBookQA}, and AristoRoBERTa \citep{clark2019f} is used for \textit{OpenBookQA} as an additional setting for fair comparison. We experiment to compare GSC with existing GNN-based QA methods, including RN \citep{santoro2017simple}, GconAttn \citep{wang2019improving}, RGCN \citep{schlichtkrull2018modeling},  KagNet \citep{kagnet-emnlp19}, MHGRN \citep{feng2020scalable} and QA-GNN \citep{yasunaga2021qagnn}, which only differ in the design of GNN reasoning modules. We report the performance of baselines referring to \citet{yasunaga2021qagnn} and all the test results are evaluated on the best model on the dev split.

\subsection{Results}

\input{obqa_main_table}

\footnotetext{See Appendix~\ref{sec:appendix_b} for the detail of this result.}

\myparagraph{\textit{CommonsenseQA}.} As shown in \tbl{tab:csqa_main}, \modelshort outperforms the previous best model with $2.57\%$ mean accuracy on In-house dev split and $1.07\%$ mean accuracy on the in-house test split. We observe that the performance variance of GSC is smaller than the baselines, indicating that our \modelshort are both effective and stable. On the official leaderboard of \textit{CommonsenseQA} in \tbl{tab:csqa_leaderboard}, \modelshort also outperforms all the GNN-based QA systems. Note that the previous top system UnifiedQA (11B params) uses T5 \citep{2020t5} as the pre-trained LM model, which is 30x larger than our model and uses much more pre-training data.

\myparagraph{\textit{OpenBookQA}.} From \tbl{tab:obqa_main} our \modelshort outperforms the previous best model with $2.53\%$ mean test accuracy with normal setting and $3.9\%$ test accuracy with the AristoRoBERTa setting. More remarkably, as shown in \tbl{tab:obqa_leaderboard}, our \modelshort ranks top one on the official leaderboard of \textit{OpenBookQA}, which even surpasses the performance of UnifiedQA (11B), which is 30x larger than our model.

\subsection{Discussion}
\input{obqa_leaderboard_table}
As mentioned above, we observe that the initial node embeddings are dispensable. Then we start to explore how the maximum number of retrieved nodes (also related to edges) affects the model. We experiment various numbers of nodes for our \modelshort, and summarize the results in \tbl{tab:ablation}. We find that larger maximum number of node does not benefit the model, and the model achieves the best performance when we use all and only the entity nodes directly occur in question and answer, whose number of nodes is generally less than 32. This indicates that 1-hop retrieval is adequate for our methods, and this could be done super efficiently than multi-hop retrieval.

\input{venn_figure}

To further analyze the reasoning capacity of \modelshort, we draw the Venn diagram for the predictions of different methods and ground truth (GT) in \fig{fig:venn}. We find that even for the different runs of the same GSC model, the correct overlap is only 69\%, showing that the datasets are relatively noisy and there exists decent variance in the prediction results. We also observe that \modelshort has a larger overlap for GNN-based systems (e.g., QA-GNN, MHGRN), while at the same time having less overlap for non-GNN methods. The ALBERT model has the least overlap since all the other methods use RoBERTa as the text encoder.

\input{ablation_table}

We also observe that the order of the percentage of overlap of left four exactly matches the order of the performance of each model: GSC > QA-GNN > MHGRN > RoBERTa. This indicates that \modelshort has quite similar behaviors versus other GNN-based systems, and the reasoning capability of \modelshort is on par with existing GNN counterparts. This further reveals that counting plays an essential role in knowledge-aware reasoning for QA.

To verify our observations, we handcraft a hard edge counting feature feeding into a simple two-layer MLP. As shown in \tbl{tab:ablation}, this hard counting model with 2-hop edge feature could achieve a comparable performance of \modelshort, which even outperforms other GNN baselines. These impressive results not only show that our \modelshort is effective, but also prove that counting is an essential functionality in the process of the current knowledge-aware QA systems.

As shown in \fig{fig:interp}, for the retrieved sub-graph of each answer choice, we can directly observe the behavior of the \modelshort by printing out the output edge/node values (soft count) of each layer. In this way, we can trace back to see why the model scores the answers like that.  We list the runtime soft counts of \modelshort's edge encoder in Appendix~\ref{sec:appendix_c}. A higher soft count means that the edge/node is more important, and can contribute more to the final graph score. This demonstrates the advantage of \modelshort as an interpretable reasoning module.

\input{interpret_figure}

%% file: csqa_main_table.tex
\begin{wraptable}{r}{0.55\textwidth}
\scalebox{0.78}{\parbox{\linewidth}{
\begin{tabular}{lcc}
    \toprule  
    {\textbf{Methods}}& \textbf{IHdev-Acc.} (\%) & \textbf{IHtest-Acc.} (\%) \\
    \midrule  
    RoBERTa-large (w/o KG)  & 73.07~($\pm$0.45) & 68.69 ($\pm$0.56) \\
    \midrule
    \ \ + {RGCN \scalebox{0.7}{\citep{schlichtkrull2018modeling}}} & 72.69~($\pm$0.19) & 68.41~($\pm$0.66) \\
    \ \ + {GconAttn \scalebox{0.7}{\citep{wang2019improving}}}  & 72.61($~\pm$0.39) &
    68.59~($\pm$0.96)\\
    \ \ + {KagNet \scalebox{0.7}{\citep{kagnet-emnlp19}}}  & 73.47~($\pm$0.22) & 69.01~($\pm$0.76) \\
    \ \ + {RN} \scalebox{0.7}{\citep{santoro2017simple}}  & 74.57~($\pm$0.91) & 69.08~($\pm$0.21) \\
    \ \ + {MHGRN} \scalebox{0.7}{\citep{feng2020scalable}} & 74.45~($\pm$0.10)   & {71.11}~($\pm$0.81)  \\
    \ \ + QAGNN \scalebox{0.7}{\citep{yasunaga2021qagnn}} &  76.54~($\pm$0.21)   & 73.41~($\pm$0.92)  \\
    \midrule

    \ \ + \modelshort (Ours) &  \textbf{79.11}~($\pm$0.22)   & \textbf{74.48}~($\pm$0.41)  \\
    
    \bottomrule 
\end{tabular}
}}
\vspace{-4pt}
\caption{Performance comparison on \textit{Commonsense \!QA} in-house split (controlled experiments). 
As the official test is hidden, here we report the in-house dev (IHdev) and test (IHtest) accuracy, following the data split of \citet{kagnet-emnlp19}.
}
\vspace{-4pt}
\label{tab:csqa_main}
\end{wraptable}

%% file: csqa_leaderboard_table.tex
\begin{wraptable}{r}{0.5\textwidth}
\vspace{-10pt}
\centering
\small
\scalebox{0.9}{
\begin{tabular}{lc}
\toprule
\textbf{Methods}& \textbf{Test}  \\
\midrule
RoBERTa~\citep{liu2019jingfei} & 72.1 \\

RoBERTa + FreeLB~\citep{zhu2019freelb} (ensemble) &73.1\\
RoBERTa + HyKAS~\citep{ma-etal-2019-towards}&73.2\\
RoBERTa + KE (ensemble) & 73.3\\
RoBERTa + KEDGN (ensemble) & 74.4\\
XLNet + GraphReason~\citep{lv2020graph} & 75.3\\
RoBERTa + MHGRN \citep{feng2020scalable} & 75.4  \\
ALBERT + PG \citep{wang2020connecting} & 75.6  \\
RoBERTa + QA-GNN \citep{yasunaga2021qagnn} & 76.1 \\
ALBERT~\citep{lan2019albert} (ensemble) &76.5          \\
UnifiedQA (11B)\textsuperscript{*}~\citep{2020unifiedqa} & \textbf{79.1} \\
\midrule
RoBERTa + \modelshort (Ours) & 76.2 \\
\bottomrule
\end{tabular}
}
\vspace{-4pt}
\caption{Test accuracy on \textit{CommonsenseQA}'s official leaderboard. The previous top system, UnifiedQA (11B params) is 30x larger than our model.
}
\vspace{-16pt}
\label{tab:csqa_leaderboard}
\end{wraptable}

%% file: obqa_main_table.tex
\begin{wraptable}{r}{0.55\textwidth}
\vspace{-10pt}
\scalebox{0.78}{\parbox{\linewidth}{
\begin{tabular}{lcc}
\toprule
\textbf{Methods}          & \textbf{RoBERTa-large}     & \textbf{AristoRoBERTa}     \\
\midrule
Fine-tuned LMs (w/o KG)          &  64.80~($\pm$2.37)  & 78.40~($\pm$1.64)       \\
\midrule
\ \ + RGCN           & 62.45~($\pm$1.57)   & 74.60~($\pm$2.53)
       \\
\ \ + GconAtten      & 64.75~($\pm$1.48)   & 71.80~($\pm$1.21)
       \\
\ \ + RN             & 65.20~($\pm$1.18)    & 75.35~($\pm$1.39)
       \\
\ \ + MHGRN & 66.85~($\pm$1.19) & 80.6        \\
\ \ + QAGNN          & 67.80\footnotemark~($\pm$2.75)   &  82.77~($\pm$1.56)    \\
\midrule
\ \ + \modelshort (Ours) &  \textbf{70.33}~($\pm$0.81)   &  \textbf{86.67}~($\pm$0.46)    \\

\bottomrule
\end{tabular}
}}

\vspace{-8pt}
\caption{Test accuracy on \textit{OpenBook \!\!QA}. Methods with AristoRoBERTa use the textual evidence by \citet{clark2019f} as an additional input to the QA context.  }
\vspace{-8pt}
\label{tab:obqa_main}
\end{wraptable}

%% file: obqa_leaderboard_table.tex
\begin{wraptable}{r}{0.5\textwidth}
\centering
\small
\scalebox{0.9}{
\begin{tabular}{lc}
\toprule
\textbf{Methods}         & \textbf{Test}           \\
\midrule
Careful Selection~\citep{banerjee2019careful} & 72.0\\
AristoRoBERTa & 77.8\\
KF + SIR~\citep{banerjee2020knowledge} & 80.0\\
AristoRoBERTa + PG \citep{wang2020connecting}      & 80.2 \\
AristoRoBERTa + MHGRN \citep{feng2020scalable}     & 80.6 \\
ALBERT + KB & 81.0\\
AristoRoBERTa + QA-GNN  & 82.8\\
T5\textsuperscript{*}~\citep{2020t5} & 83.2\\
UnifiedQA(11B)\textsuperscript{*}~\citep{2020unifiedqa} & 87.2         \\
\midrule
AristoRoBERTa + GSC (Ours) & $\textbf{87.4}$ \\
\bottomrule
\end{tabular}
}
\caption{Test accuracy on \textit{OpenBookQA} leaderboard. All listed methods use the provided science facts as an additional input to the language context. The previous top 2 systems, UnifiedQA (11B params) and T5 (3B params) are  30x and 8x larger than our model.}
\label{tab:obqa_leaderboard}
\end{wraptable}

%% file: venn_figure.tex
\begin{figure}[t]
    \centering
    \vspace{-14pt}
    \includegraphics[width=\linewidth]{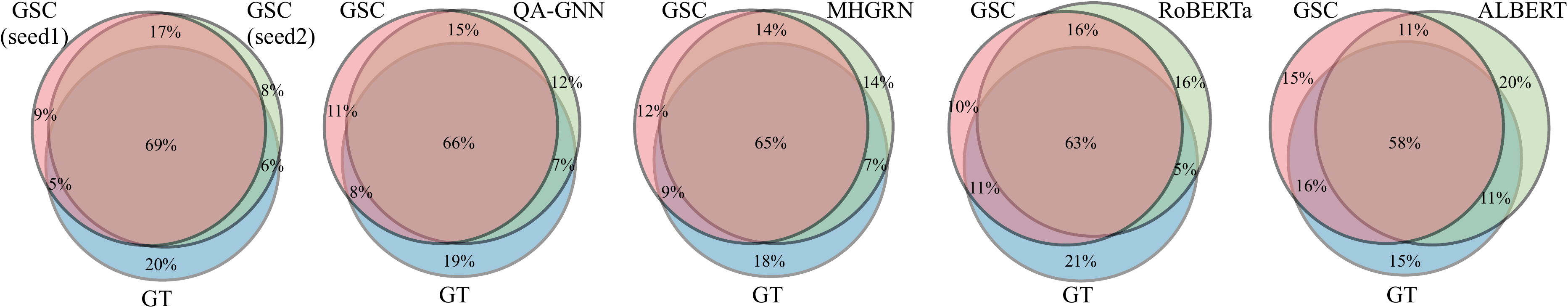}
    \vspace{-14pt}
    \caption{Venn diagrams for the prediction overlap of different models and ground truth (GT) of the IHtest split of CommonsenseQA. The ALBERT has the least overlap since the all left four use RoBERTa as text encoder. The order of the percentage of overlap of left four  exactly matches the order of the performance of each models: GSC > QA-GNN > MHGRN > RoBERTa. This indicates that the reasoning ability learned by our \modelshort basically covers that of other GNNs. }
    \label{fig:venn}
    \vspace{-14pt}
\end{figure}

%% file: ablation_table.tex
\begin{wraptable}{r}{0.55\textwidth}
\vspace{-8pt}
\scalebox{0.85}{\parbox{\linewidth}{
\begin{tabular}{lcc}
    \toprule  
    {\textbf{Methods}}& \textbf{IHdev-Acc.} (\%) & \textbf{IHtest-Acc.} (\%) \\
    \midrule  
    MLP + Counter \scalebox{0.8}{(1-hop)}   & 78.02~($\pm$0.05) & 73.62~($\pm$0.12) \\
    MLP + Counter \scalebox{0.8}{(2-hop)}   & 78.30~($\pm$0.09) & 74.13~($\pm$0.08) \\
  
    \midrule
    \modelshort w/~\scalebox{0.8}{QA nodes} &  79.11~($\pm$0.22)   & 74.48~($\pm$0.41)  \\
    \midrule
    ~~~~~~~~~w/ ~\scalebox{0.8}{32 nodes} &  78.52~($\pm$0.58)   & 74.40~($\pm$0.25)  \\ 
    ~~~~~~~~~w/ ~\scalebox{0.8}{64 nodes} &  78.53~($\pm$0.77)   & 73.93~($\pm$1.09)  \\ 
    ~~~~~~~~~w/~\scalebox{0.8}{128 nodes} &  78.50~($\pm$0.96)   & 72.78~($\pm$1.15)  \\ 
    ~~~~~~~~~w/~\scalebox{0.8}{256 nodes} &  78.32~($\pm$0.60)   & 73.89~($\pm$0.63)  \\
    \bottomrule 
\end{tabular}
}}
\vspace{-4pt}
\caption{Ablation study on the hard counter with MLP (upper) and the maximum number of retrieved nodes (bottom), showing that 1) the hard counter achieves performance on par with GNN-based methods; 2) GSC works well even when only using entities occurred in QA context, which typically contains less than 32 nodes.}

\vspace{-10pt}
\label{tab:ablation}
\end{wraptable}

%% file: interpret_figure.tex
\begin{figure}[t]
    \vspace{-14pt}
    \centering
    \includegraphics[width=0.95\linewidth]{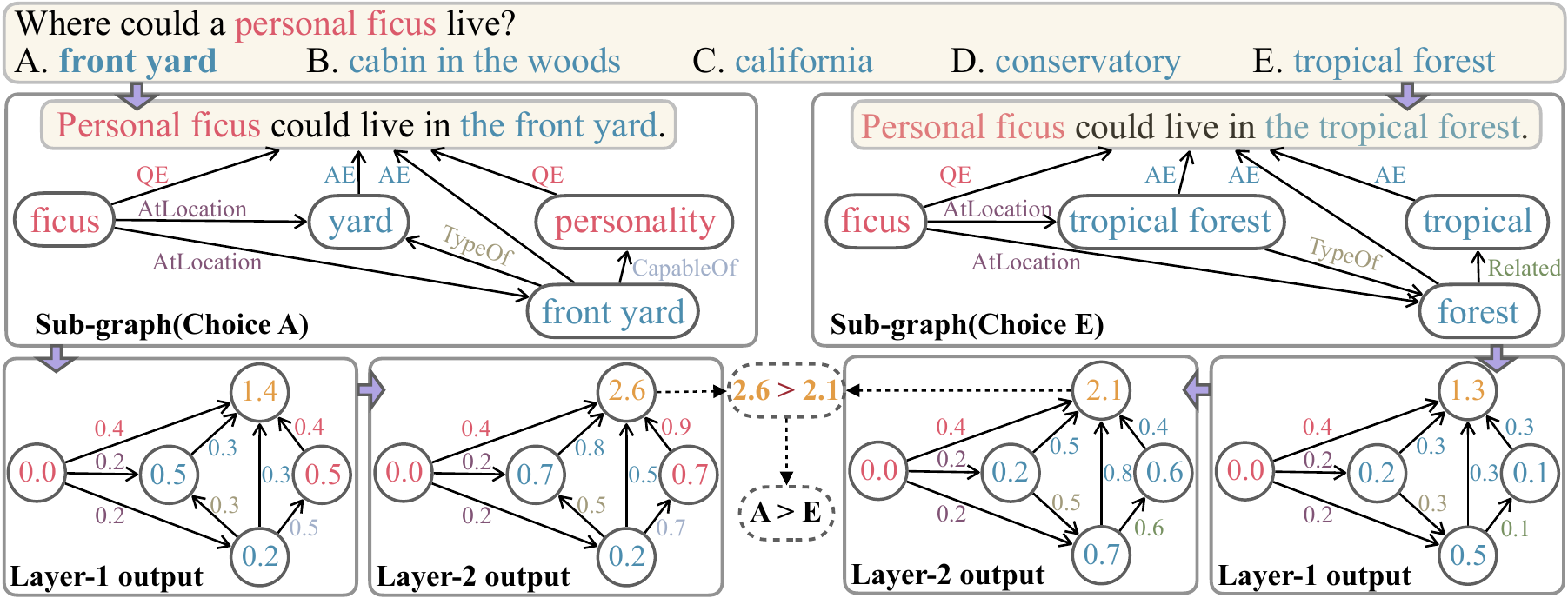}
    \caption{Our \modelshort is highly interpretable. For the retrieved sub-graph of each answer choice, we can directly observe the behaviour of the model by print out the output edge/node values of each layer, so that we can trace back to see how the model score the answer.}
    \label{fig:interp}
    \vspace{-14pt}

\end{figure}

%% file: 5_related.tex
\section{Related Work}
\label{sec:related}

\myparagraph{KG-powered QA.}
For KG-powered QA, traditional methods use semantic parsers to map the question to logical form~\citep{BerChoFro13,YihChaHe15} or executable program on KG~\citep{LiaBerLeFor16}, which typically require domain-specific rules and fine-grained annotations, making them difficult to handle the noise in questions and KG. To address these challenges, various deep neural models have been applied to KG-powered QA, such as memory networks~\citep{Kumaretal15,SukWesFeretal115,MilFisDodKaretal16}, neural programmer for knowledge table QA~\citep{NeeLeAbaMccAmo16}, neural knowledge-aware reader~\citep{xiong-etal-2019-improving,sun-etal-2019-pullnet}, etc.

\myparagraph{GNN for KG-powered QA.}
To further improve the neural reasoning capability, recent studies have explored applying GNNs to KG-powered QA, where GNNs naturally fit the graph-structured knowledge and show prominent results. KagNet \citep{kagnet-emnlp19} proposes GCN-LSTM-HPA for path-based relational graph representation. MHGRN \citep{feng2020scalable} extend Relation Networks \citep{santoro2017simple} to multi-hop relation scope and unifies both path-based models and RGCN \cite{schlichtkrull2018modeling} to enhance the interpretability and the scalability. QA-GNN \citep{yasunaga2021qagnn} proposes a LM+GAT framework to joint reasoning over language and KG. We do not use conventional GNNs and our \modelshort is extremely simple and efficient even without parameters inside the \modelshort layers. KG embeddings are generally used in these systems. \citet{zhang2018variational} employs propagation of KG embeddings to perform multi-hop reasoning. KagNet pre-trains KG Embedding using TransE initialized with GloVe embeddings. Recently, MHGRN and QA-GNN are proposed, which leverage pre-trained LM to generate embeddings for KG. In addition, QA-GNN also scores KG node relevance using LM as extra node embedding. We follow the data pre-processing procedure of QA-GNN and MHGRN, but our model does not use any KG embedding or node score embedding. 

%% file: 6_conclusion.tex
\section{Conclusion}
\label{sec:conclusion}
We investigate state-of-the-art GNN-based QA systems, and discover that they are over-parameterized. Our diagnostic analysis using SparseVD shows that the initial node embeddings and some GNN layers are completely dispensable. Inspired by our observations, we design a much simpler yet effective counter-based model named \model (\modelshort). Surprisingly, with less than 1\% trainable parameters, our model outperforms state-of-the-art GNN counterparts on two popular QA benchmark datasets. Our work reveals that counting plays a crucial role in the knowledge-aware reasoning process. It remains as a challenging open problem to build more comprehensive human-level reasoning modules for QA.

%% file: 7_appendix.tex
\section{Appendix}
\label{sec:appendix}

\subsection{Sparse Ratio Curve}
\label{sec:appendix_a}
We attach the details of the training curve of SparseVD dissection for QA-GNN as following. We find the linear query and key layers inside GNN layers and graph pooler generally can be compressed to a relatively low sparse ratio and this is more obvious at the fist three GNN layers. This indicated the these layers may be over-parameterized and the attention mechanism \citep{vaswani2017attention, velivckovic2017graph} may degenerate to linear function in this case if we remove these layers.

\input{svd_curve_detail_figure}

As expected, the final fully connected layer (FC) preserved the highest sparse ratio, since it serves for directly generating the overall score of a choice and it is very efficient (the output dimension is 1). Except that, the edge encoder with a two-layer MLP has very high sparse ratio, which can be explained as the edge encoding is essential for GNN reasoning in these tasks. This is also why we only keep the edge encoder with relatively more learnable parameters and simplify the other components in GNN to design our \model.

\subsection{OpenBookQA Results}
\label{sec:appendix_b}
We find a small error on the Table 4 of QA-GNN paper \citep{yasunaga2021qagnn}.
According to the principle that the test results should be evaluated on the best model on the dev split, instead of directly picking the best model on test split. So we compute the result of QA-GNN on OpenBookQA dataset refer to the official worksheets\footnote{\url{https://worksheets.codalab.org/worksheets/0xf215deb05edf44a2ac353c711f52a25f}} released by the author. We find this result is different from the result reported on the paper: We report [seed0: 70.20, seed1: 64.80, seed2: 68.40, mean: 67.80, std: 2.75] as the test results which are evaluated the best model on the dev split,  while the original paper report [seed0: 70.20, seed1: 69.60, seed2: 71.80, mean: 70.53, std: 1.14] as the test results which are directly picking the best model on test split. We mention this minor difference of the reported numbers here in case someone may be confused by it. And QA-GNN is a excellent method to solve QA tasks with GNN and LM.

\subsection{Soft Count of Edges}
\label{sec:appendix_c}
We list the top-30 edge triplets with highest soft counts that encoded by the edge encoder of our \modelshort as following. These combinations of edge types and node types have relatively higher counts, which means they can contribute more to the final graph score.

\input{edge_soft_count_table}

\subsection{SparseVD}
\label{sec:appendix_d}

To take a deep scope of the graph neural networks based knowledge-aware systems, we introduce a neural model pruning method Sparse Variational Dropout (SparseVD,   \citep{molchanov2017variational}) into this scenario as a dissection tool to automatically dissect the graph network architecture.  We implemented the SparseVD optimization based layer refer to the PyTorch code\footnote{\url{https://colab.research.google.com/github/bayesgroup/deepbayes-2019/blob/master/seminars/day6/SparseVD-solution.ipynb}} released by the author. To keep the dissection in strict accordance with the theoretical derivation of \citet{molchanov2017variational}, we apply regulation to all the linear layers in the GNN modules. We attach the SparseVD problem formulation as well as the optimization target as following.

Consider a dataset $\mathcal{D}$ which is constructed from $N$ pairs of objects $(x_n, y_n)_{n=1}^N$.
Our goal is to tune the parameters $w$ of a model $p(y\cond x, w)$ that predicts $y$ given $x$ and $w$.
In Bayesian Learning we usually have some prior knowledge about weights $w$, which is expressed in terms of a prior distribution $p(w)$.
After data $\mathcal{D}$ arrives, this prior distribution is transformed into a posterior distribution $p(w\cond \mathcal{D})=p(\mathcal{D}\cond w)p(w)/p(\mathcal{D})$.
This process is called \emph{Bayesian Inference}.
Computing posterior distribution using the Bayes rule usually involves computation of intractable multidimensional integrals, so we need to use approximation techniques.

One of such techniques is \emph{Variational Inference}.
In this approach the posterior distribution $p(w\cond\mathcal{D})$ is approximated by a parametric distribution $q_\phi(w)$.
The quality of this approximation is measured in terms of the Kullback-Leibler divergence $D_{KL}(q_\phi(w)\,\|\,p(w\cond \mathcal{D}))$.
The optimal value of variational parameters $\phi$ can be found by maximization of the \emph{variational lower bound}:
\begin{equation}
   \mathcal{L}(\phi) =  L_\mathcal{D}(\phi) - D_{KL}(q_\phi(w)\,\|\,p(w)) \to\max_{\phi\in\Phi}
   \label{eq:elbo}
\end{equation}
\vspace{-1em}
\begin{equation}
   L_\mathcal{D}(\phi) = \sum_{n=1}^N \mean_{q_\phi(w)}[\log p(y_n\cond x_n, w)]
   \label{lD}
\end{equation}
It consists of two parts, the expected log-likelihood $L_\mathcal{D}(\phi)$ and the KL-divergence $D_{KL}(q_\phi(w)\,\|\,p(w))$, which acts as a regularization term.

We formulate the model prunings (parsification) problem  pruning problem as a variational inference problem and use the Kullback-Leibler Divergence approximation following \citep{achterhold2018variational, molchanov2017variational} to constrain the model parameters to converge to the pruning prior:
Original $-D_{KL}$ was obtained by averaging over $10^7$ samples of $\epsilon$ with less than $2\times10^{-3}$ variance of the estimation.
\begin{equation}
\begin{gathered}
    -D_{KL}(q(w_{ij}\cond\theta_{ij}, \alpha_{ij})\,\|\,p(w_{ij})) \approx
    \\
    \approx k_1\sigma(k_2 + k_3\log \alpha_{ij})) - 0.5\log(1+\alpha_{ij}^{-1}) + \mathrm{C}
    \label{eq:KL}\\
    k_1=0.63576~~~~~k_2=1.87320~~~~~k_3=1.48695
\end{gathered}
\end{equation} 
Where $\sigma(\cdot)$ denotes the sigmoid function. We can see that $-D_{KL}$ term increases with the growth of $\alpha$. It means that this regularization term favors large values of $\alpha$. The case of $\alpha_{ij}\to \infty$ corresponds to a Binary Dropout rate $p_{ij}\to 1$ (recall $\alpha=\frac{p}{1-p}$). Intuitively it means that the corresponding weight is almost always dropped from the model. Therefore its value does not influence the model during the training phase and is put to zero during the testing phase.

We can also look at this situation from another angle.
Infinitely large $\alpha_{ij}$ corresponds to infinitely large multiplicative noise in $w_{ij}$.
It means that the value of this weight will be completely random and its magnitude will be unbounded.
It will corrupt the model prediction and decrease the expected log likelihood. 
Therefore it is beneficial to put the corresponding weight $\theta_{ij}$ to zero in such a way that $\alpha_{ij}\theta_{ij}^2$ goes to zero as well.
It means that $q(w_{ij}\cond\theta_{ij}, \alpha_{ij})$ is effectively a delta function, centered at zero $\delta(w_{ij})$.
\begin{equation}
\begin{gathered}
    \theta_{ij}\to 0,\ \ \ \  \alpha_{ij}\theta_{ij}^2\to0\\
    \Downarrow\\
    q(w_{ij}\cond\theta_{ij},\alpha_{ij})\to\Normal(w_{ij}\cond0, 0)=\delta(w_{ij})
\end{gathered}
\end{equation}
In the case of linear regression this fact can be shown analytically.
We denote a data matrix as $X^{N\times D}$ and  $\alpha,\theta\in\RR^D$.
If $\alpha$ is fixed, the optimal value of $\theta$ can also be obtained in a closed form.
\begin{equation}
\label{eq:vdrvr}
\theta = (X^\top X + \diag(X^\top X)\diag(\alpha))^{-1} X^\top y
\end{equation}
Assume that $(X^\top X)_{ii} \neq 0$, so that $i$-th feature is not a constant zero.
Then from (\ref{eq:vdrvr}) it follows that $\theta_{i}=\Theta(\alpha_{i}^{-1})$ when $\alpha_{i}\to+\infty$, so both $\theta_{i}$ and $\alpha_{i}\theta_{i}^2$ tend to $0$.

In our dissection scenario, we optimize the weight $\theta_{ij}$to zero no longer for compressing the model but for figuring out which part of the model can be pruned out (sparse ratio to zero) without loss of accuracy, which indicates that part of model is not important and its output information is unused in optimization.

%% file: svd_curve_detail_figure.tex
\begin{figure*}[h]
    \centering
    \includegraphics[width=\linewidth]{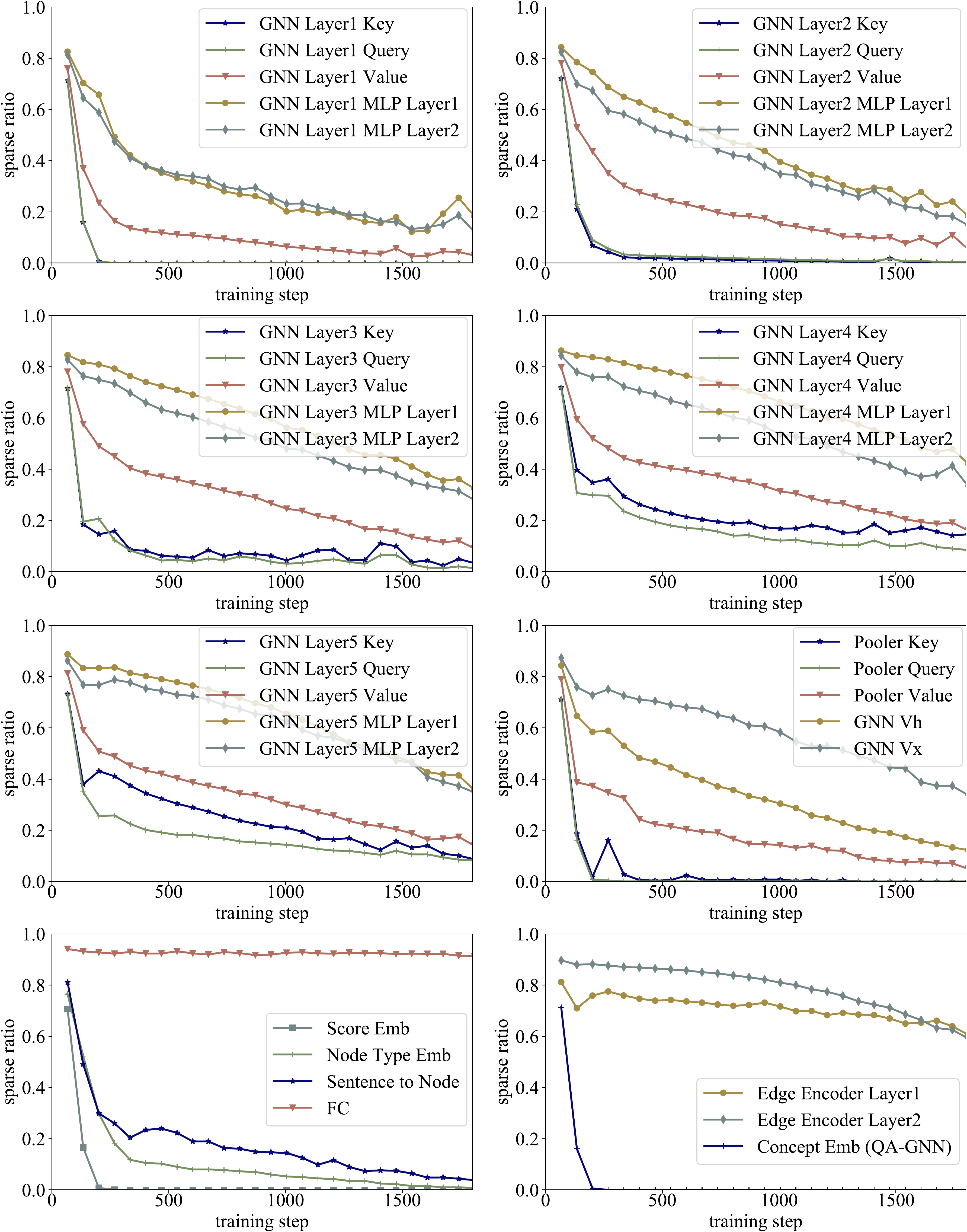}
    \caption{The sparse ratio curve when do SparseVD training for the QA-GNN systems.}
    \label{fig:svd_curve_detail}
\end{figure*}

%% file: edge_soft_count_table.tex
\begin{table}[t]
\renewcommand*{\arraystretch}{0.1}
\centering

\scalebox{1.06}{\parbox{\linewidth}{
\begin{tabular}{c|l|c|c}
    \toprule  
    \textbf{Head Node Type} & \textbf{Relation Type} & \textbf{Head Node Type} & \textbf{Soft Count Value}   \\
\midrule
Q entity & desires (inverse) & Q entity & 0.9632  \\ 
\midrule
Q entity & desires & Q entity & 0.8902  \\ 
\midrule
Q entity & has subevent (inverse) & Q entity & 0.8775  \\ 
\midrule
Q entity & is & Q entity & 0.8470  \\ 
\midrule
Q entity & has property & Q entity & 0.7344  \\ 
\midrule
Q entity & does not desires & Q entity & 0.6995  \\ 
\midrule
Q entity & does not desires (inverse) & Q entity & 0.6790  \\ 
\midrule
Q entity & causes & Q entity & 0.6252  \\ 
\midrule
Q entity & has property (inverse) & Q entity & 0.3531  \\ 
\midrule
A entity & is & Q entity & 0.3073  \\ 
\midrule
Q entity & causes (inverse) & Q entity & 0.2868  \\ 
\midrule
Q entity & is part of & Q entity & 0.2758  \\ 
\midrule
Q entity & is capable of (inverse) & Q entity & 0.2706  \\ 
\midrule
A entity & has property & Q entity & 0.2424  \\ 
\midrule
A entity & desires & Q entity & 0.2207  \\ 
\midrule
Q entity & is the antonym of (inverse) & Q entity & 0.2137  \\ 
\midrule
Q entity & is created by (inverse) & Q entity & 0.1720  \\ 
\midrule
Q entity & is capable of & Q entity & 0.1681  \\ 
\midrule
Q entity & is part of (inverse) & Q entity & 0.1581  \\ 
\midrule
Q entity & desires (inverse) & A entity & 0.1310  \\ 
\midrule
Q entity & is at location of & Q entity & 0.1006  \\ 
\midrule
Q entity & is created by & Q entity & 0.0990  \\ 
\midrule
Q entity & is at location of (inverse) & Q entity & 0.0914  \\ 
\midrule
Q entity & is & A entity & 0.0737  \\ 
\midrule
Q entity & has property & A entity & 0.0636  \\ 
\midrule
Q entity & has subevent (inverse) & A entity & 0.0617  \\ 
\midrule
O entity & is & Q entity & 0.0606  \\ 
\midrule
O entity & has property & Q entity & 0.0512  \\ 
\midrule
Q entity & desires & A entity & 0.0445  \\ 
\midrule
Q entity & is the antonym of & Q entity & 0.0399  \\ 
    \bottomrule 
\end{tabular}
}}

\caption{We list the top-30 edge triplets with highest soft counts here, and it is generated by the edge encoder of \modelshort model. The combination of edge types and node types with a higher count means it can contribute more to the final graph score. }

\label{tab:edge_soft_count}
\end{table}